# HSR: $L_{1/2}$ Regularized Sparse Representation for Fast Face Recognition using Hierarchical Feature Selection

Bo Han[a], Bo He[a,*], Tingting Sun[a], Mengmeng Ma[a], Amaury Lendasse[b]

[a]*School of Information and Engineering, Ocean University of China,*

*Shandong, Qingdao, China 266000*

[b]*Department of Mechanical and Industrial Engineering and the Iowa Informatics*

*Initiative, 3131 Seamans Center, The University of Iowa,*

*Iowa City, IA 52242-1527, USA*

**Abstract**

In this paper, we propose a novel method for fast face recognition called $L_{1/2}$ Regularized Sparse Representation using Hierarchical Feature Selection (HSR). By employing hierarchical feature selection, we can compress the scale and dimension of global dictionary, which directly contributes to the decrease of computational cost in sparse representation that our approach is strongly rooted in. It consists of Gabor wavelets and Extreme Learning Machine Auto-Encoder (ELM-AE) hierarchically. For Gabor wavelets part, local features can be extracted at multiple scales and orientations to form Gabor-feature based image, which in turn improves the recognition rate. Besides, in the presence of occluded face image, the scale of Gabor-feature based global dictionary can be compressed accordingly because redundancies exist in Gabor-feature based occlusion dictionary. For ELM-AE part, the dimension of Gabor-feature based global dictionary can be compressed because high-dimensional face images can be rapidly represented by low-dimensional feature. By introducing $L_{1/2}$ regularization, our approach can produce sparser and more robust representation compared to $L_1$ regularized Sparse Representation based Classification (SRC), which also contributes to the decrease of the computational cost in sparse representation. In comparison with related work such as SRC and Gabor-feature based SRC (GSRC), experimental results on a variety of face databases demonstrate the great advantage of our method for computational cost. Moreover, we also achieve approximate or even better recognition rate.

# 1 Introduction

The technique of face recognition plays an important role in people's life ranging from commercial to law enforcement applications, such as real-time surveillance, biometric personal identification, and information security[1]. It is one of the most challenging topics in the interface of computer vision and cognitive science. Over past 40 years, extensive research on face recognition has been conducted by many psychophysicists, neuroscientists and engineers. In general views, the definition of face recognition can be formulated as follows. Different faces in a static image can be identified using a database of stored faces. Available collateral information like facial expression may enhance the recognition rate. Generally speaking, if the face images are sufficiently provided, the quality of face recognition will be mainly related to feature extraction and recognition modeling.

For feature extraction, more specifically, there are roughly two kinds of popular face features including holistic features and local features. However, the classical methods using holistic features such as Eigenface[2], Fisherface[3] and Randomface are hardly to reveal the essential structures of high-dimensional faces[4]. Therefore, researchers recently prefer local-feature based methods like subspace learning[5] or manifold representation[6]. On one hand, high-dimensional images can be effectively projected into low-dimensional subspace or sub-manifold. On the other hand, compared to holistic-feature based approaches, local-feature based approaches are always less sensitive to variations of illumination, viewpoint and expression, which in turn improves the recognition rate.

For recognition modeling, lots of researchers usually evaluate the performance of model by recognition rate instead of computational cost. Recently, Wright and Ma[7] reported their work called the sparse representation based classification (SRC). To be more specific, it can represent the testing image sparsely using training samples via $L_1$-norm minimization, which can be solved by balancing the minimum reconstructed error and the sparsest coefficients. Experimental results showed that the recognition rate of SRC is much higher than that of classical algorithms such as Nearest Neighbor, Nearest Subspace and Linear Support Vector Machine (SVM). However, there are three drawbacks behind the SRC. First, SRC is based on the holistic features, which cannot exactly capture the partial deformation of the face images. Second, $L_1$ regularized SRC usually runs slowly for high-dimensional face images. Third, in the presence of occluded face images, Wright et al. introduce an occlusion dictionary to sparsely code the occluded components in face images. However, the computational cost of SRC increase drastically because of large number of elements in the occlusion dictionary. Therefore, the computational cost of SRC limits its application in real-time area, which increasingly attracts researchers' attention to solve this issue.

Recently, Yang and Zhang's work (Gabor-feature based SRC (GSRC)) [8] claimed that if Gabor wavelets[9] can be employed in feature extraction, it is possible to obtain a much more compact occlusion dictionary in the presence of occluded faces, which not only speeds up the computation but also improves the recognition rate. Although the GSRC provides us a good insight about how to reduce the computational cost of SRC in the presence of occluded faces, to our best knowledge, there is still an essence issue to be addressed. Namely, the computational cost of sparse representation is highly related to three aspects including the dimension of face images, the scale of occlusion dictionary and the speed of regularized optimization. Without any occlusion, Gabor wavelets mainly play an important role in local features extraction. So the computational cost of SRC can be determined by the dimension of face images and $L_1$-norm minimization instead of the scale of occlusion dictionary. If we want to reduce computational cost of SRC in a general condition, on the one hand, we should effectively project high-dimensional faces into low-dimensional features. On the other hand, we should find a sparser representation than $L_1$ regularized SRC.

Inspired by these observations, in this paper, we propose a novel method for fast face recognition called $L_{1/2}$ Regularized Sparse Representation using Hierarchical Feature Selection (HSR). In the feature extraction, we employ hierarchical feature selection because it contributes to the decrease of computational cost in sparse representation that our approach is strongly rooted in. It consists of Gabor wavelets and Extreme Learning Machine Auto-Encoder (ELM-AE)[10] hierarchically. To be more specific, Gabor wavelets could effectively extract local features at multiple scales and orientations[11] forming Gabor-feature based images, which can greatly improve the recognition rate. Moreover, in the presence of occluded faces, we can obtain a compact occlusion dictionary via sparse coding because of redundancies in Gabor-feature based occlusion dictionary, thus the scale of global dictionary can be decreased accordingly. In addition, high-dimensional face images can be effectively represented by low-dimensional features via ELM-AE, thus the dimension of global dictionary can be decreased accordingly. So for Gabor-feature based global dictionary, the compression of scale and dimension contributes to the decrease of computational cost in sparse representation. Finally, the Gabor-feature based methods have been applied into face recognition leading to state-of-the-art recognition rate[12]. Also the computational cost of ELM-AE is much less than Principal Component Analysis (PCA) used in SRC. In the recognition modeling, the main difference between our method HSR and SRC is that $L_1$-norm minimization is replaced by $L_{1/2}$-norm minimization[13] because $L_{1/2}$-norm minimization can produce sparser representation, which directly decreases the computational cost of sparse representation. Although $L_{1/2}$-norm minimization belongs to non-convex optimization problems, it can be easily transformed into a series of weighted $L_1$-norm minimization, which is convenient for us to solve by existing methods. Moreover, $L_{1/2}$-norm minimization is more robust than $L_1$-norm minimization, which is more suitable to process occluded face images. In our experiments, the new method has been verified on representative face databases (Extended Yale B, AR and FERET) with different conditions like lighting, pose, expression, and occlusion. In comparison with related work such as SRC[7] and GSRC[8], experimental results demonstrated that our method is slightly complicated in structure, but it shows the great advantage for computational cost. And we also achieve approximate or even better recognition rate. Therefore, our method has a great potential for the application of fast face recognition like real-time surveillance.

The rest of paper is organized as follows. In section 2, we briefly discuss previous work on ELM and sparse representation based on $L_1$ regularization. In section 3, we describe our new method including hierarchical feature selection and $L_{1/2}$ regularized sparse representation. In section 4, we report experimental results on SRC, GSRC and our method HSR under representative face databases with different conditions. Also we present discussions on the performance of new method. Finally, in section 5, we show conclusions on our current research and indicate two important directions for future work.

## 2 Previous works
### 2.1 The structure of the original ELM

Extreme Learning Machine (ELM) was proposed by Huang et.al for faster learning speed and higher generalization performance[14][15]. The essence of ELM is that the parameters of the hidden nodes can be generated randomly without manually tuning[16]. Specifically speaking, the input data $x$ is mapped to $L$-dimensional hidden layer and the network output is given by Eq(1).

$$f_L(x) = \sum_{j=1}^{L} \beta_j h_j(x) \tag{1}$$

Where $\beta_j = [\beta_{j1}, \beta_{j2}, ..., \beta_{jm}]^T$ are the output weights between the hidden nodes and the output nodes, $h_j(x) = g_j(x^T \alpha_j + b_j)$ is the output of hidden layer, $\alpha_j$ is the input weight, $b_j$ is the input bias and $g_j(x)$ is the activation function, they all correspond with the output of the $jth$ hidden node. The ELM algorithm can be summarized as follows. Given $N$ training samples $\{x_i, t_i\}_{i=1}^{N}$, where input data $x_i = [x_{i1}, x_{i2}, ..., x_{in}]^T$ and the target labels $t_i = [t_{i1}, t_{i2}, ..., t_{im}]^T$. The input data is mapped to $L$-dimensional hidden layer initially. The structure of ELM can be determined if the output weights $\beta$ can be calculated, so the following learning problems can be formulated by Eq(2).

$$H\beta = T \tag{2}$$

Where $T = [t_1, t_2, ..., t_N]^T$ is the target matrix, $H = [h(x_1), ..., h(x_i), ..., h(x_N)]$ is the hidden layer output matrix, and $h(x_i) = [h_1(x_i), h_2(x_i), ..., h_L(x_i)]^T$. So the output weights $\beta$ can be calculated by Eq(3).

$$\beta = H^\dagger T \tag{3}$$

Where $H^\dagger$[15][17] denotes the Moore-Penrose generalized inverse of matrix $H$.

To make the resultant solution more stable and have better generalization performance[10], a positive value $I/C$ as a regularization term can be added to the diagonal of $H^T H$ shown Eq(4).

$$\beta = \left(\frac{I}{C} + H^T H\right)^{-1} H^T H \tag{4}$$

### 2.2 Sparse representation based on $L_1$ regularization

Given training samples $A_i = [s_{i,1}, s_{i,2}, ..., s_{i,n_i}] \in \mathbb{R}^{m \times n_i}$ from all the $ith$ training samples, where $s_{i,j}(j = 1,2, ..., n_i)$ is an $m$-dimensional vector, which belongs to the $jth$ sample of the $ith$ class. Denote by $y_0 \in \mathbb{R}^m$, a test sample from the same $ith$ class. Intuitively, $y_0$ can be approximately represented by the linear combination of the training samples within $A_i$.

$$y_0 = \alpha_{i,1} s_{i,1} + \alpha_{i,2} s_{i,2} + \cdots + \alpha_{i,n_i} s_{i,n_i} = \sum_{j=1}^{n_i} \alpha_{i,j} s_{i,j} \tag{5}$$

Suppose that the test sample $y_1$ is initially unknown of the exact class, a new matrix $A$ is defined to concatenate the entire training samples of all $k$ classes:

$$A = [A_1, A_2, \ldots, A_k] = [s_{1,1}, s_{1,2}, \ldots, s_{k,n_k}] \qquad (6)$$

Then the linear representation of $y_1$ can be naturally written as Eq(7).

$$y_1 = A\alpha \qquad (7)$$

According to sparse coding via $L_1$-norm minimization, the sparse coefficients $\alpha$ can be calculated as Eq(8).

$$\hat{\alpha}_1 = \arg\min_{\alpha}\{\|y_1 - A\alpha\|_2^2 + \lambda\|\alpha\|_1\} \qquad (8)$$

In the case of occluded data, we should express the test sample $y_2$ as a sum of sparse representation and error. Then the previous model[7] can be modified as Eq(9).

$$y_2 = A\alpha + e_0 = [A, A_e]\begin{bmatrix}\alpha\\ \alpha_e\end{bmatrix} = B\omega \qquad (9)$$

Where $B \in \mathbb{R}^{m \times (n_A + n_{A_e})}$, and the $e_0 \in \mathbb{R}^m$ is a noise term with bounded energy $\|e_0\|_2 < \varepsilon$. According to sparse coding via $L_1$-norm minimization, the sparse coefficients $\omega$ can be calculated as Eq(10).

$$\hat{\omega}_1 = \arg\min_{\omega}\{\|y_2 - B\omega\|_2^2 + \lambda\|\omega\|_1\} \qquad (10)$$

Therefore, we can represent test sample $(y_1, y_2)$ as sparse coefficients $(\hat{\alpha}_1, \hat{\omega}_1)$, which can be employed to identify the class of test sample.

## 3  $L_{1/2}$ Regularized Sparse Representation using Hierarchical Feature Selection

### 3.1 Framework

In this section, we briefly introduce the new method called $L_{1/2}$ Regularized Sparse Representation using Hierarchical Feature Selection (HSR). Our method will tackle with two critical issues in face recognition. First, how can we reduce computational cost of recognition modeling while keeping the recognition rate? Second, how can we ensure the robustness of our method to occluded faces? Our approach roughly consists of feature extraction and recognition modeling, which solves above problems accordingly. And also we provide the convincing reasons for the choice of recognition model and parameters. The structure of HSR is shown in Fig1.

For feature extraction, by employing hierarchical feature selection, we can compress the scale and dimension of global dictionary, which directly contributes to the decrease of computational cost in sparse representation that our approach is strongly rooted in. To be more specific, it consists of Gabor wavelets and ELM-AE hierarchically. For Gabor wavelets part, according to theories of visual neuroscience, the mechanism of retina cells in human eyes can be simply simulated by Gabor wavelets, which could effectively extract local features at multiple scales and orientations. And the local-feature based methods are always less sensitive to variations of illumination, viewpoint and expression. Therefore, Gabor-feature based

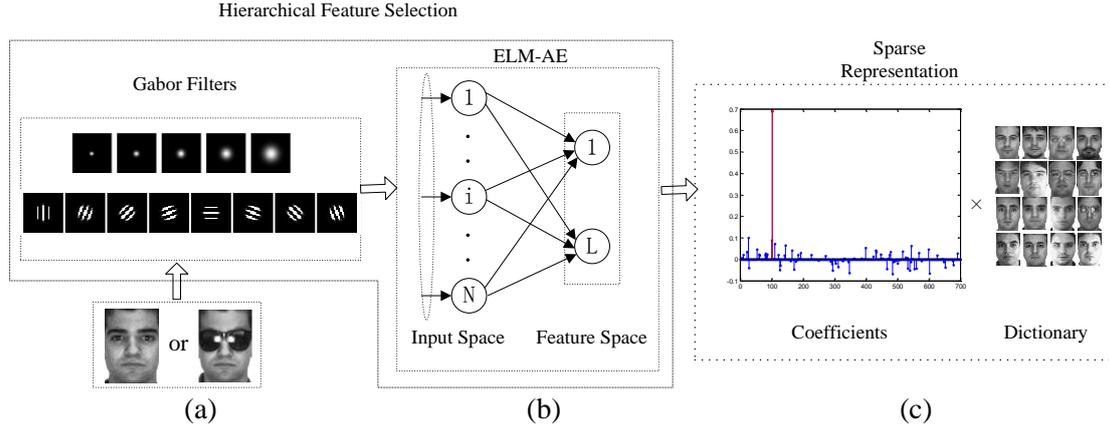

Fig1. The framework of HSR

images can improve recognition rate to some extent. Moreover, for the occluded images, the enormous scale of occlusion dictionary is the principle factor to affect the computational cost of sparse representation. Because of redundancies exist in Gabor-feature based occlusion dictionary, the scale of Gabor-feature based global dictionary can be compressed. Besides, the Gabor-feature based methods have been applied into face recognition leading to state-of-the-art recognition rate like Liu and Wechsler's work. For ELM-AE part, we hope to modify the basic ELM to represent input training and testing images meaningfully. Namely, the output weight of ELM-AE is responsible of learning the features from the input data via singular values. According to ELM theory, ELM-AE is a universal approximator that has a strong ability to achieve compressed, sparse, and equal dimension representation. So it is reasonable to believe that images in a higher dimensional input space can be effectively projected into a lower dimensional feature space via ELM-AE. Thus the dimension of Gabor-feature based global dictionary can be compressed. Moreover, the computational cost of ELM-AE is much less than that of PCA used in SRC because of its random weights and biases of the hidden nodes.

For recognition modeling, more specifically, our approach is strongly rooted in the framework of sparse representation, which has showed its excellent performance on recognition rate especially for occluded faces. The testing image can be sparsely represented by the linear combination of the training samples and our target is to balance the reconstructed error and the sparsest coefficients via different kinds of regularization. In our approach, we choose $L_{1/2}$-norm minimization instead of $L_1$-norm minimization used in SRC or another regularized parameters because of two reasons. First, $L_1$ regularization locates between $L_0$ regularization and $L_\infty$ regularization, so $L_1$ regularization has sparse property and it can be solved easily. Naturally thinking, $L_{1/2}$ regularization locates between $L_0$ regularization and $L_1$ regularization, so we expect that $L_{1/2}$ regularization has sparser property than $L_1$ regularization. Actually, the geometry property of $L_{1/2}$ and $L_1$ regularization has obviously proved our expectation. Second, Xu's experiments[13] demonstrated that the performance of sparse representation using $L_{1/2}$ regularization is stronger than that using other $L_p$ regularization ($0<p<1/2$ or $1/2<p<1$). One might argue that $L_{1/2}$-norm minimization belongs to non-convex optimization problems, which means it is hard to solve. However, we can transform it into a series of weighted $L_1$-norm minimization, which is convenient for us to solve by existing methods. Moreover, according to Xu's experiments, $L_{1/2}$-norm minimization is more robust than $L_1$-norm minimization, which is more suitable to process occluded faces.

### 3.2 Hierarchical Feature Selection

In sparse representation, the compression of global dictionary normally comes from the reduction of dimension and scale (the number of elements), which directly contributes to the decrease of computational cost. For hierarchical feature selection, we employ Gabor wavelets and ELM-AE hierarchically. By employing Gabor wavelets, we can initially represent original images by Gabor-feature based images, which can improve the recognition rate. In addition, when the face images are partially occluded, we can compress the scale of Gabor-feature based occlusion dictionary via sparse coding, thus the scale of Gabor-feature based global dictionary can be compressed accordingly. By using ELM-AE, high-dimensional images can be rapidly represented by low-dimensional features, thus we can compress the dimension of Gabor-feature based global dictionary and testing images.

#### 3.2.1 Gabor-Feature based Image Representation and Occlusion Dictionary

The motivation that we choose Gabor wavelets for image representation is mainly due to their biological relevance and computational properties. In this section, we will mainly formulate how to represent original image via Gabor wavelets below. Then, in the presence of occluded images, we briefly introduce how to compress the scale of Gabor-feature based occlusion dictionary via sparse coding.

More specifically, Gabor wavelets[12] usually demonstrate good characteristics of spatial locality and orientation selectivity. Moreover, in the space and frequency domains, they are optimally localized. It can also be defined with the orientation $\mu$ and scale $\nu$ as follows.

$$\psi_{\mu,\nu}(z) = \frac{\|k_{\mu,\nu}\|}{\sigma^2} e^{\left(-\|k_{\mu,\nu}\|^2 \|z\|/2\sigma^2\right)} \left[e^{ik_{\mu,\nu}z} - e^{-\sigma^2/2}\right] \quad (11)$$

Where the pixel of an image is $z = (x, y)$, the wave vector is defined as $k_{\mu,\nu} = k_\nu e^{i\phi_\mu}$ with $k_\nu = k_{max}/f^\nu$ and $\phi_\mu = \pi\mu/8$. $k_{max}$ is the maximum frequency, and $f$ is the spacing factor between kernels in the frequency domain. Besides, $\sigma$ determines the ratio of the Gaussian window width to wavelength. In most cases, Gabor wavelets have five different scales and eight orientations. As Liu and Wechsler's work, the real part of Gabor wavelets can be shown in Fig2.(a).

Here we should also note that when the parameters of Gabor wavelets are as $\sigma = 2\pi$, $k_{max} = \pi/2$, $f = \sqrt{2}$, the Gabor wavelets demonstrate the excellent characteristics of spatial frequency, orientation selectivity and spatial locality.

According to above discussion, we can naturally represent high-dimensional images via Gabor wavelets. The Gabor-feature based local representation is equal to the convolution of the input images with each Gabor wavelet. For example, the convolution of image $img(z)$ with a Gabor wavelet is defined as below.

$$G_{\mu,\nu}(z) = img(z) * \psi_{\mu,\nu}(z) \quad (12)$$

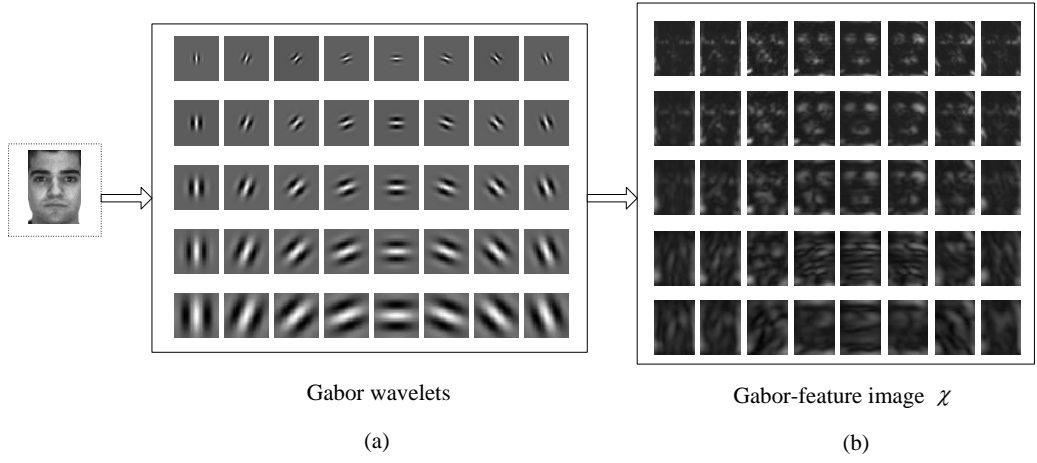

Fig2 (a)The real part of the Gabor wavelets at five scales and eight orientations, (b)Gabor-feature based image $\chi$

These convolution results show different scales, localities and orientations corresponding to the Gabor wavelets. As Liu and Wechsler's work, the convolution results are all complex number. To contain a Gabor-feature based image, we should first normalize all convolution results, and then concatenate them to form an augmented feature vector $\chi$.

$$\chi = \left(G_{0,0}^T G_{0,1}^T \ldots G_{4,7}^T\right)^T \tag{13}$$

Where $\chi$ is an image based on Gabor-feature, which not only improves recognition rate but also bears to image local deformation to some degree. $\chi$ can be shown in Fig2.(b).

To make a further step, we can derive Gabor-feature based Image Representationin two situations. First, without any occlusion, the linear representation of $y_1$ can be rewritten as $y_1 = A\alpha$. By employing Gabor wavelets, we can derive the Eq(14).

$$\chi(y_1) = X(A_1)\alpha_1 + X(A_2)\alpha_2 + \cdots + X(A_k)\alpha_k = X(A)\alpha \tag{14}$$

Where $X(A) = [X(A_1), X(A_2), \ldots, X(A_k)]$ and $X(A_i) = [\chi(s_{i,1}), \ldots, \chi(s_{i,n_i})]$.

In the presence of occluded testing image $y_2$, the Eq(14) should be modified as Ma's work indicated as Eq(15).

$$\chi(y_2) = [X(A), X(A_e)] \begin{bmatrix} \alpha \\ \alpha_e \end{bmatrix} = X(B)\omega \tag{15}$$

Where $X(A_e)$ is the Gabor-feature based occlusion dictionary, and $\alpha_e$ is the representation coefficient vector of the input Gabor feature vector $\chi(y_2)$ over $X(A_e)$.

The occlusion dictionary $A_e$ in SRC is normally selected as the identity matrix $I$[7], so SRC has a large number of elements in occlusion dictionary, which definitely increases the computational cost of optimization. For example, If the dimension of original images is $83 \times 60(4980)$, then the dimension of occlusion dictionary $I$ in SRC is of $4980 \times 4980$. By using Gabor wavelets, the occlusion dictionary will be represented into Gabor-feature based occlusion dictionary $(4980 \times 40 \times 4980)$, which is obviously redundant. So we should compress it from two aspects including the dimension and scale. Now, let's briefly discuss the scale compression. We hope to compress the scale of Gabor-feature based occlusion dictionary because redundancies exist in it. For example, suppose $z = X(A_e) = [z_1, ..., z_{n_e}] \in \mathbb{R}^{m_p \times n_e}$ the original Gabor-feature based occlusion dictionary, then the scale-compressed occlusion dictionary is denoted by $\Gamma = [d_1, ..., d_p] \in \mathbb{R}^{m_p \times p}(p \ll n_e)$, and we can represent $Z$ by $\Gamma$ via sparse coding. So our objective function is defined as Eq(16).

$$J_{\Gamma,\Lambda} = arg\ min\{\|Z - \Gamma\Lambda\|_F^2 + \varsigma\|\Lambda\|_1\} \quad s.t. d_j^T d_j = 1\ \forall j \tag{16}$$

It is easy to solve this optimization problem by optimizing $\Gamma$ and $\Lambda$ alternatively. Therefore, the compression of scale is easily achieved. The Eq(15) can be modified by Eq(17)

$$\chi(y_2) = [X(A), \Gamma]\begin{bmatrix}\alpha \\ \alpha_\Gamma\end{bmatrix} \tag{17}$$

In the next section, we will introduce a new method called ELM-AE, which can effectively compress the dimension of global dictionary.

### 3.2.2 ELM-AE for High-Dimensional Images Representation

The motivation that we choose ELM-AE for image representation is due to its representation ability and computational cost. We will mainly introduce ELM-AE for high-dimensional images representation below. And then we will briefly verify the performance of ELM-AE.

For Auto-Encoder[18], the output data $\hat{x}$ is similar to the input data $x$. Some interesting structure can be obtained when constraints are placing on the networks. For example, suppose that the input image is $10 \times 10$, and the number of the input nodes is 100, but only 50 hidden nodes, then the network must try to reconstruct 100-dimension output nodes with the 50 hidden nodes, which is forced to learn a compressed representation a compressed representation of the input. Based on the above concept, the ELM-AE was first proposed by Huang et.al[10], and the main objective of ELM-AE is to represent the input data meaningfully and rapidly. In term of the number of the input nodes and hidden nodes, there are three different representation including compressed representation, sparse representation and equal dimension representation. For face recognition task, we hope to represent input training and testing images by compressed representation. The training structure of ELM-AE can be seen as Fig3.(a).

To be more specific, we first modify the basic ELM[15][19] to conduct unsupervised learning $(t = x)$, and random weights and biases of the hidden nodes are chosen to be orthogonal because orthogonalization will make the generalization of ELM-AE better.

The orthogonal random weight and bias can be calculated by Eq(18).

$$a^T a = I, b^T b = 1 \tag{18}$$

Where $a = [a_1, ..., a_L]$ is the orthogonal random weight and $b = [b_1, ..., b_L]$ is the orthogonal random bias between the input nodes and hidden nodes. Then we calculate the hidden layer output matrix $H$ as the original ELM does.

$$H = g(ax + b) \tag{19}$$

After training process, we first hypothesize that the output weights of ELM can be treated as coding parameters of Auto-Encoder, which is responsible of learning the low-dimensional features from the high-dimensional data. And the output weight $\beta$ as Eq(20).

$$\beta = \left(\frac{I}{C} + H^T H\right)^{-1} H^T X \tag{20}$$

Where $H = [h_1, \ldots, h_N]$ is the output of hidden layer and $X = [x_1, \ldots, x_N]$ is the input data.

Therefore, the trained ELM-AE will be employed to conduct high-dimensional images representation, for example, the original image is initially represented via Gabor wavelets, and the dimension of Gabor-feature based image will naturally increase. Then we represent Gabor-feature based image $\chi$ into low-dimensional features $\chi_r$ via ELM-AE by Eq(21).

$$\chi_r = \chi \beta \tag{21}$$

After hierarchical feature selection, the image representation $\chi_r$ can be visualized in Fig3.(b), which will be sent into the framework of sparse representation for further processing.

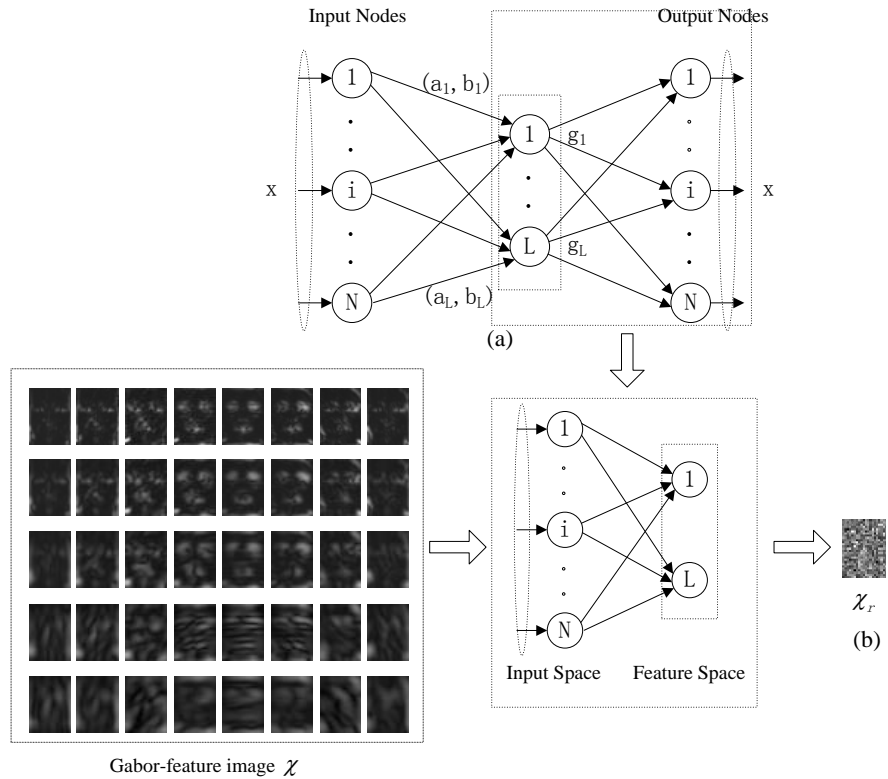

Fig3. (a)The training of ELM-AE. (When the number of hidden nodes is less than that of input nodes, it can be used as the compressed representation), (b) Image representation after hierarchical feature selection

We can clearly see that the computational cost of ELM-AE is much lower than that of PCA. It shows that ELM-AE itself can speed up the process of feature extraction. Moreover, in HSR, the ELM-AE can effectively reduce the dimension of global dictionary and testing images, which greatly relieve the computational burden undertaken by optimization methods.

### 3.3 $L_{1/2}$ Regularized Sparse Representation

For recognition modeling, our approach is strongly rooted in the framework of sparse representation because of good recognition rate and robustness to occluded faces. In this section, we first introduce generic framework of sparse representation, and then compare different regularized parameters such as $L_{1/2}$, $L_1$ and $L_2$. Finally, we decide to employ $L_{1/2}$ because it can produce sparser and more robust representation compared to $L_1$, which speeds up the face recognition.

#### 3.3.1 Generic Framework of Sparse Representation for Face Recognition

Specifically speaking, supposing that well-aligned training face images of each class are sufficiently provided. We collect training images together forming a large training dictionary and each column is normalized via $L_2$-norm. One classical assumption is that a new image of $ith$ class can be well represented as a linear combination of all the $ith$ training samples. However, if the identity of the test image is unknown, the problem becomes more complex because we will represent the test image using training dictionary $A$. So the linear representation of the test image $y_1$ can be written as Eq(22).

$$y_1 = A\alpha \in \mathbb{R}^m \quad (22)$$

Where $\alpha$ is a vector of sparse coefficients. In practical situations, the coefficient vector is often complicated because of the presence of partial occluded faces $y_2$. The linear model should be modified as Eq(23).

$$y_2 = A\alpha + e_0 = [A, A_e]\begin{bmatrix}\alpha \\ \alpha_e\end{bmatrix} = B\omega \quad (23)$$

Where $e_0 \in \mathbb{R}^m$ is a vector of error, $B = [A, A_e]$ and $\omega = [\alpha, \alpha_e]^T$, thus face recognition in the presence of occlusion can be represented as the sparest coefficients. We hope to introduce more general framework of sparse representation, so we will choose general loss function $L$ for there constructed error and an uncertain norm for regularized parameters. Therefore, the sparest coefficients can be represented as Eq(24).

$$\widehat{\omega}_k = arg\min_{\omega}\{L(y_2, B\omega) + \lambda\|\omega\|_k\} \quad (24)$$

Where $\|\cdot\|_k$ denotes $k$-norm that represents the uncertain parameter. Therefore, we can represent test sample $y_2$ using sparse coefficients $\widehat{\omega}_k$. The general framework can almost explain all special cases. For example, if general loss function $L$ can be denoted as square loss, then the framework can be converted into AIC[20] and BIC[21] criteria when $k = 0$, the framework can be converted into Lasso algorithm[22] when $k = 1$, the framework can be converted into ridge regression[23] when $k = 2$. In our approach, we choose traditional square loss function, and the next section will discuss the choice of regularized parameters.

### 3.3.2 Regularized Parameters: $L_0$, $L_{1/2}$ and $L_1$

The motivation why we choose $L_{1/2}$-norm minimization is due to two aspects.

First, although the sparsest coefficients can be obtained via $L_0$-norm minimization, the procedure of solving $L_0$-norm minimization is a NP-hard problem. Therefore, Tibshirani introduced the Lasso algorithm ($L_1$-norm minimization) to obtain relative sparse coefficients, which is much easier for us to solve because $L_1$-norm minimization belongs to the convex problem. What's more, they proved that $L_0$ regularization is equal to $L_1$ regularization on the certain constraint condition.

However, in the practical application, we find that $L_1$-norm minimization usually cannot produce the sparest solution, so a question is raised that whether we can introduce a new regularized parameter, which provides a sparser solution than $L_1$ regularization. Fortunately, Xu's experimental results proved that $L_{1/2}$ regularization can produce sparser representation compared with $L_1$ regularization, which is also proved by their geometry property. From the Fig4, the bound of $L_p$ regularization has different shape, and the solution of $L_p$-norm minimization is equal to the intersection of the bound and the loss function. For example, we can clearly see that solution via $L_2$ is not sparse at all and the solution via $L_{1/2}$ is sparser than that via $L_1$ because the bound of $L_{1/2}$ is easier to intersect with loss function at coordinates.

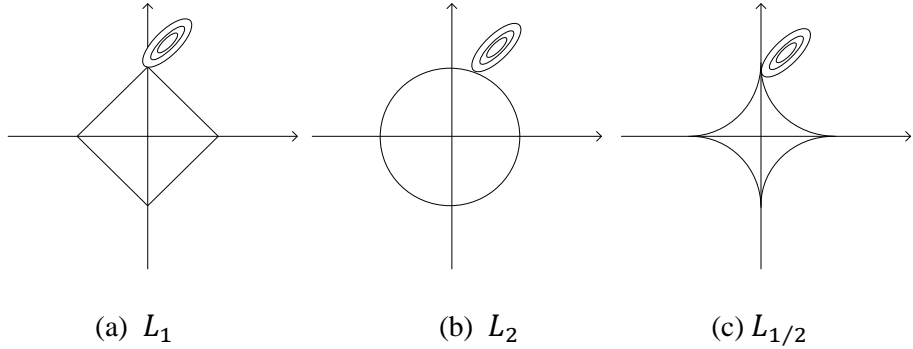

(a) $L_1$  (b) $L_2$  (c) $L_{1/2}$

Fig4. The possibility of sparse solution via $L_1$, $L_2$ and $L_{1/2}$

Although $L_{1/2}$-norm minimization belongs to non-convex optimization problems, we can transform it into a series of weighted $L_1$-norm minimization, which is also convenient for us to solve by existing methods. Moreover, according to Xu's experiments, $L_{1/2}$-norm minimization is more robust than $L_1$-norm minimization, which is more suitable to process occlusion in face images.

Second, although we initially want to explore other possibilities like $L_p$-norm minimization ($0<p<1/2$ or $1/2<p<1$), Xu's experiments[13] clearly demonstrated that the performance of sparse representation using $L_{1/2}$ regularization is stronger than that using other $L_p$ regularization ($0<p<1/2$ or $1/2<p<1$). Therefore $L_{1/2}$ regularization can completely replace $L_p$ regularization ($0<p<1$).

Overall, we naturally introduce $L_{1/2}$ regularized sparse representation for fast face recognition using hierarchical feature selection (HSR). Our method has solved two critical issues raised in the section3.1. First, we employ Gabor wavelets and ELM-AE hierarchically in order to reduce the dimension and scale of global dictionary, which accordingly reduces the computational cost of sparse representation. Second, $L_{1/2}$-norm minimization can produce sparser and more robust representation than $L_1$-norm minimization, which not only reduces the computational cost of optimization but also is more suitable to process occlusion in face images. The algorithm of HSR and its explanation are released below.

---

**Algorithm 1** The HSR algorithm

1. Images representation: $y_2$ is an occluded face, which is a special case of normal face. The columns of $A$ and $A_e$ are normalized to have unit $L_2$-norm.

$$y_2 = A\alpha + e_0 = [A, A_e]\begin{bmatrix}\alpha \\ \alpha_e\end{bmatrix} = B\omega$$

Where $y_2 \in \mathbb{R}^{m\times 1}, B \in \mathbb{R}^{m\times(n_A+n_{A_e})}$.

2. Gabor wavelets: extract local feature to enhance recognition rate and compress the scale of Gabor feature based occlusion dictionary via sparse coding.

$$\chi(y_2) = [\chi(A), \chi(A_e)]\begin{bmatrix}\alpha \\ \alpha_e\end{bmatrix} = B_\chi \omega$$

$$\chi(y_2) = [\chi(A), \Gamma]\begin{bmatrix}\alpha \\ \alpha_\Gamma\end{bmatrix} = B_\Gamma \omega_\Gamma$$

Where $\chi(y_2) \in \mathbb{R}^{m_\chi \times 1}$, $B_\chi \in \mathbb{R}^{m_\chi \times (n_A+n_{A_e})}$, $B_\Gamma \in \mathbb{R}^{m_\chi \times (n_A+n_\Gamma)}$, and $n_\Gamma \ll n_{A_e}$.

3. ELM-EA: compress the dimension of global dictionary and testing image

$$\chi_r(y_2) = [\chi_r(A), \Gamma_r]\begin{bmatrix}\alpha \\ \alpha_\Gamma\end{bmatrix} = B_r^\Gamma \omega_\Gamma$$

Where $\chi_r(y_2) \in \mathbb{R}^{m_r \times 1}$, $B_r^\Gamma \in \mathbb{R}^{m_r \times (n_A+n_\Gamma)}$, and $m_r \ll m_\chi$.

4. Solve the $l_{1/2}$-norm minimization problem: get a sparser and more robust representation.

$$\widehat{\omega}_{\Gamma_{1/2}} = arg\min_\omega\{\|\chi_r(y_2) - B_r^\Gamma \omega_\Gamma\|_2^2 + \lambda\|\omega_\Gamma\|_{1/2}^{1/2}\}$$

Where

$$\widehat{\omega}_{\Gamma_{1/2}} = [\hat{\alpha}_{1/2}, \hat{\alpha}_{\Gamma_{1/2}}]$$

$$B_r^\Gamma = [\chi_r(A), \Gamma_r]$$

And $\lambda$ is a positive scalar number that balances the reconstructed error and sparse coefficients.

5. Compute the residuals

$$r_i(\chi_r(y_2)) = \|\chi_r(y_2) - \Gamma_r\hat{\alpha}_{\Gamma_{1/2}} - \chi_r(A)\delta_i(\hat{\alpha}_{1/2})\|_2, \text{ for } i = 1,2,\dots,k.$$

Where $\delta_i(\cdot): \mathbb{R}^n \to \mathbb{R}^n$ is the characteristic function which selects the coefficients associated with the $i^{th}$ class.

6. Output that $identity(y_2) = identity(\chi_r(y_2)) = arg\min_i r_i(\chi_r(y_2))$

## 4 Experimental results

In this section, we present some experimental results on available benchmark databases to compare the performance of the proposed algorithm HSR with GSRC and SRC. The reason why HSR does not compare with deep learning algorithms[24] is due to the common fact that their computational cost is very expensive. To evaluate the performance of HSR comprehensively, this section is divided into two detailed sections. In section 4.1 we first tested our method on the face datasets without occlusion. And then in section 4.2 we tested the new method on the face datasets against occlusion using two different frameworks (no partition and partition). All the simulations for the HSR, GSRC and SRC algorithms are carried out in Matlab 7.8 environment running in an Intel Xeon E5-1650 3.20GHz CPU. In the experiments of Gabor wavelets, the parameters are set as $k_{max} = \pi/2$, $f = \sqrt{2}$, $\sigma = \pi$, eight orientations $\mu = \{0, ..., 7\}$ and five different scales $v = \{0, ..., 4\}$ by our experience. And the parameters are fixed for all the experiments below. The activation function of ELM is set to 'sig' representing the sigmoidal function, the parameter $C$ is set to 100 and the number of the hidden neurons is equal to the compressive feature space dimension. In addition, all face images provided in the databases are cropped and aligned by the location of eyes. The face images from the databases are further normalized to zero mean and unit variance.

### 4.1 Face recognition without occlusion

We compared the performance of the HSR with two classical algorithms SRC and GSRC on three typical facial image databases: Extended Yale B[25], AR[26] and FERET[27]. For the Extended Yale B and AR databases, we compared the performance of HSR, SRC and GSRC versus feature dimension. Moreover, we compared the performance of different compression methods (ELM-AE and PCA) and different regularized parameters ($L_1$ and $L_{1/2}$) on two databases. For the FERET, we compared the performance of HSR, SRC and GSRC versus pose angle.

1) Extended Yale B Database: The database consists of 2414 frontal-face images of 38 individuals. The images are normalized to 96×84 under various laboratory-controlled lighting conditions. We randomly selected half of the images for training (i.e., 32 images per subject), and the other half for testing. Choosing the training set randomly assures that our results will be independent of any special choice. Fig5 shows some samples from the same object class, and it is obvious that only illumination is added to these images. The dimension of the Gabor-feature based image is 199200(83×60×40) through a set of Gabor wavelets, which includes five different scales and eight orientations. They can capture abundant local features to form Gabor-feature based image, which will take a lot of time to process this high-dimensional image. To compress the feature space, we applied ELM-AE (a part of HSR) and PCA (a part of SRC and GSRC) respectively with the feature dimensions 30, 56, 120, 224 and 504 on the Gabor-feature based images. Then we computed the recognition rate and the computational cost. In addition, the computational cost of sparse representation is equal to the testing time because there is no training process. In our experiments, we set $\lambda = 0.001$[7] in HSR, GSRC and SRC by our experience. It shows the recognition rates in Fig6.(a) and computational cost in Fig6.(b) of HSR comparing with GSRC and SRC versus the feature dimension. It is turned out that with the increase of feature dimension, the recognition rate becomes higher and the computational cost becomes more. HSR achieves a maximum recognition rate of

98.52% with 504D feature space. In contrast, the maximum recognition rate of GSRC is 98.44% and SRC is 97.50%. The computational cost of SRC and new method is similar, which much less than that of GSRC. According to a specific dimension $504D$, the computational cost of compression by PCA is $36.2022s$ while ELM-AE is $1.3011s$.

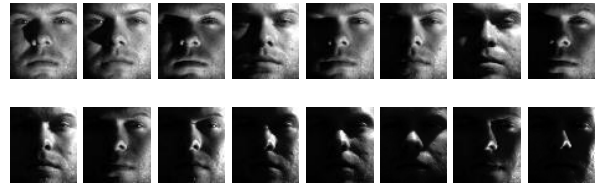

Fig5.Samples from the same object of Extended Yale B dataset

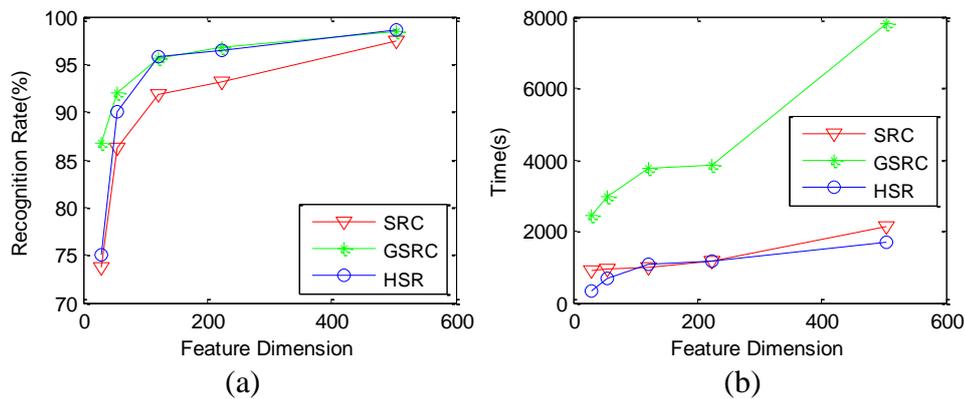

Fig6. Recognition rates (a) and time (b) by SRC, GSRC and HSR versus feature dimension across Extended Yale B database

2) AR Database: The AR database consists of 4000 frontal images from 126 individuals. We chose a subset consisting of 50 male subjects and 50 female subjects. For each subject, 14 images are selected, which includes only illumination changes and expressions. Fig7 shows several samples from the same object class with the variation of expression and illumination. We selected seven images from Session 1 for training and seven images from Session 2 for testing. The images were cropped and converted to gray scale with the size is $83 \times 60$. The dimension of the Gabor-feature vector is 12000 after a set of Gabor wavelets. Then we continued to reduce the feature space with five dimensions: 30, 54, 130, 311 and 540. We also set $\lambda = 0.001$ in HSR, GSRC and SRC, like on the Extended Yale B database. it shows the recognition rates in Fig8.(a) and the computational cost in Fig8.(b) of HSR comparing with GSRC and SRC versus the feature dimension. On this database, the maximum recognition rate of HSR, GSRC and SRC are 95.86%, 95.86% and 93.57% respectively. The computational cost of HSR is much less than the other two methods on different dimensions. To be more specific, the total computational time of HSR is about 49% less than that of SRC, and about 77% less than that of GSRC.

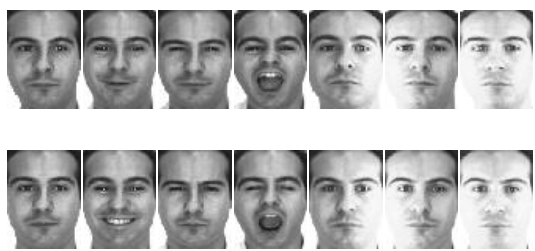

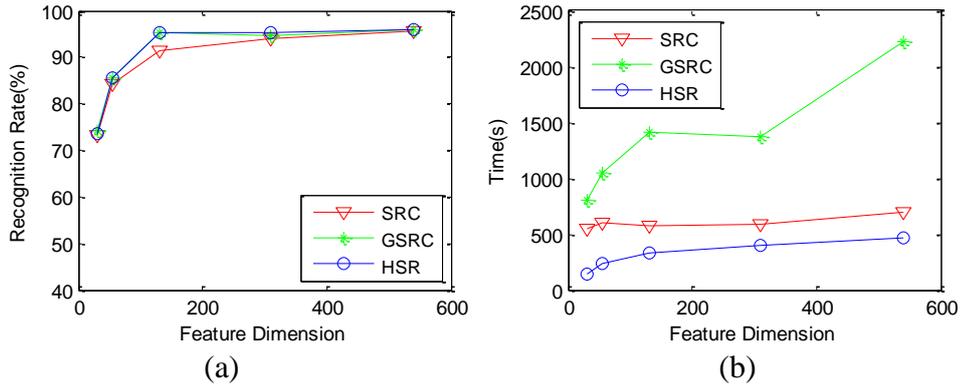

Fig7.samples from the same object of AR dataset

(a)  (b)

Fig8. Recognition rates (a) and time (b) by SRC, GSRC and HSR versus feature dimension across AR database

On the whole, the AR database is more challenging than the Extended Yale B database, thus the total recognition rate of the AR database is declining slightly. This is due to the AR database includes 100 subjects, but the training samples (dictionary atoms) are only seven images. With more stringent conditions, different lighting conditions are added into four neutral faces and different expressions are added into three faces per subject for AR database. In contrast, for the Extended Yale B database, the number of each subject is larger, and the scale of global dictionary is bigger. Only illumination variations exist on the images. The above two experiments illustrate the performance of HSR is much better than that of the SRC and GSRC versus feature dimension especially for computational cost.

For briefly verifying the compression performance of ELM-AE, we selected PCA as a control group for high-dimensional images representation. Only the computational cost of the compression component was taken into consideration in our experiments. We compared the computational cost of ELM-AE and PCA on the Extended Yale B and the AR database before the process of recognition modeling. The testing sets of the Extended Yale B and the AR are compressed to a certain dimension (405D for the Extended Yale B and 450D for the AR), whose computational costs are list on the follow Table 1.

Table1 : Comparation on computational cost of ELM-AE and PCA

|  | Databases | Yale | AR |
|---|---|---|---|
| Time(s) | ELM-AE | 1.3011 | 0.4596 |
|  | PCA | 36.2022 | 9.8460 |

We also conducted a quantitative experiment using different compression methods (ELM-AE and PCA) and different regularized parameters ($L_1$-norm and $L_{1/2}$-norm) on two databases (table 2). For one thing, when testing sets are compressed to a certain dimension (405D for the Extended Yale B and 450D for the AR), we demonstrated that $L_{1/2}$-norm minimization is superior to the $L_1$-norm minimization for computational cost while keeping the approximate recognition rate. For another thing, ELM-AE and PCA are used to compress the Gabor-feature based images in different mechanisms. So under the same regularized parameter, the computational cost of methods using ELM-AE is less.

Table 2: experiments on different dimension compression methods (ELM-AE and PCA) and different minimization frame ($L_1$ and $L_{1/2}$) of the sparse problem across the two databases

| Database | | Yale | | AR | |
|---|---|---|---|---|---|
| methods | | ELM-AE | PCA | ELM-AE | PCA |
| $L_1$ | Rec.rate(%) | 97.12 | 98.44 | 95.71 | 95.86 |
| | Time(s) | 4117.0 | 7786.6 | 909.4263 | 2227.2 |
| $L_{1/2}$ | Rec.rate(%) | 98.52 | 98.27 | 95.86 | 95.00 |
| | Time(s) | 1708.3 | 2937.4 | 465.0887 | 729.29 |

(3) FERET pose database: this database includes 1400 images from 200 subjects (7 images per subject). Among the 1400 images, 600 images are the frontal face with illumination and facial expressions and the others are the face variation with different pose angles. The images marked with 'ba','bd', 'be', 'bf', 'bg', 'bj' and 'bk' stand for the different illumination, facial expressions and pose angles (Fig9). In our experiments, the images of this database were already cropped to the size of $80×80$. In order to examine the robustness of HSR comparing with the other original algorithms, we tested the recognition rates and computational cost with respect to the variable pose angle. Then in the first test, we used images marked with 'ba' and 'bj' for training, images marked with 'bk' for testing. In another four tests, images marked with 'ba', 'bj' and 'bk' were used as training set, and the rest of images were respectively used as testing set. After feature extraction, the dimension was fixed on 350D in above three methods. We set the parameters $\lambda = 0.005$ for HSR and GSRC and $\lambda = 0.05$ for SRC, which will conduct the best results. The results showed a growing trend of the recognition rate with less pose angle variability in Fig10.(a). When the pose angle becomes larger, the recognition rate of HSR is almost 40% higher than the nearest competitor but still poor. Besides, the computational cost of HSR and GSRC is much less than that of SRC in Fig10 (b). The above experiment illustrated the performance of HSR is also much better than that of the SRC and GSRC versus pose angle.

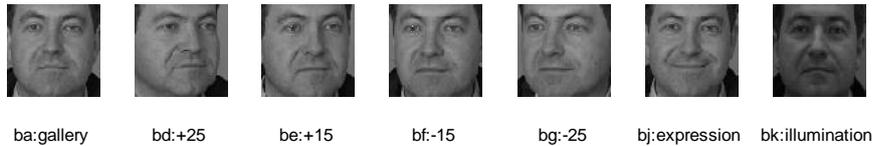

ba:gallery    bd:+25    be:+15    bf:-15    bg:-25    bj:expression    bk:illumination

Fig9. Samples from the same object class of FERET database

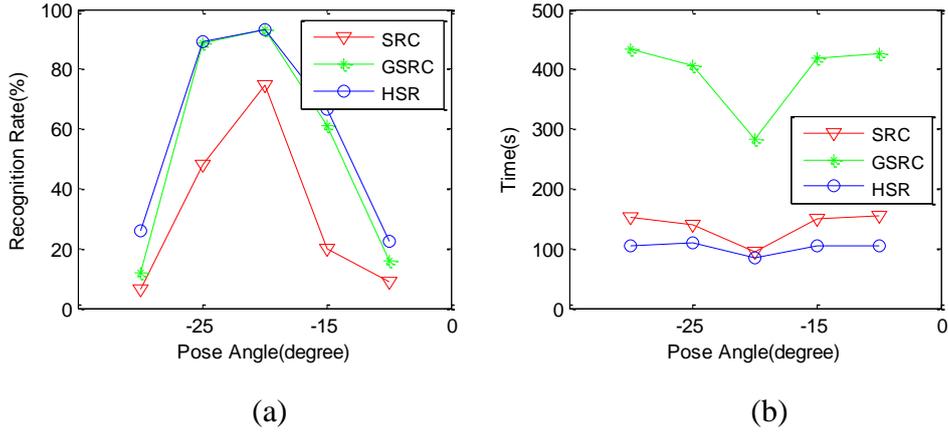

(a)                  (b)

Fig10. Recognition rates (a) and time (b) by SRC, GSRC and HSR versus pose angle across FERET database

### 4.2 Face recognition with occlusion

In this section, we also compared the performance of the HSR with SRC, GSRC on a subset of AR dataset, which includes occluded images. The chosen subset consists of 1300 images from 100 subjects (50 male and 50 female). In this subset, 700 images of unoccluded frontal face with expression and illumination variation were used for training set. Besides, the rest data were split into two separate test sets of equal size. The first test set contains 300 images, on which all the 100 subjects are wearing sunglasses. The second test set also contains 300 images, and all the subjects wear scarves instead. Sunglass occludes about 20% of the image and scarf occludes about 40% of the image intuitively (Fig11).

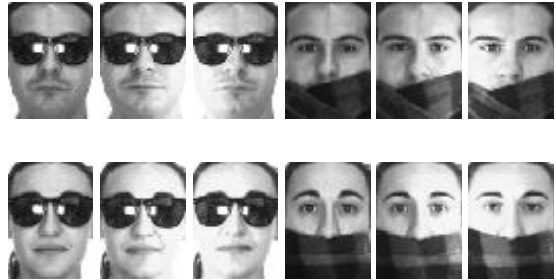

Fig11: samples of occluded faces with sunglasses and scarves on AR database

The parameters of HSR and GSRC were set to $\lambda = 0.0005$ and SRC used $\lambda = 0.005$, which will conduct the best results. The images of this dataset were resized to $83 \times 60$, then the size of global dictionary is $4980 \times 5680$ in the original SRC. In the case of the proposed HSR and GSRC, the dimension of Gabor-feature based image is $83 \times 60 \times 40$, and then decreases to 5600D by ELM-AE. Meanwhile, the scale of Gabor-feature based occlusion dictionary is compressed to 100 by sparse coding. As a result, after hierarchical feature selection, the size of global dictionary is $5600 \times 800$. Table 3 has shown the experimental results on two testing sets implemented by SRC, GSRC and HSR. Apparently, SRC performs the worst recognition rate and the highest computational cost, in other words, the holistic features used in SRC are not suitable for the occluded images and the scale of global dictionary decides the computational cost of sparse representation to some degree. Besides, it is clearly seen that the computational cost of HSR is much less than that of GSRC in two datasets while the recognition rate of HSR is higher than that of GSRC in AR scarves.

Table 3: performances of non-partitioning methods (SRC, GSRC and HSR)

| Testing set | AR sunglasses | | | AR scarves | | |
|---|---|---|---|---|---|---|
| methods | SRC | GSRC | HSR | SRC | GSRC | HSR |
| Rec.rate(%) | 79.69 | 94.33 | 85.67 | 49.00 | 90.67 | 93.67 |
| Time(s) | 13402.00 | 4788.30 | 1826.90 | 137610.00 | 4782.00 | 1935.10 |

We quoted the approach in [Wright][7] to partition the whole image into blocks and processed each block independently, assuming the occlusion part is contiguous. In these blocks, some of them are assumed to be completely occluded and some of them may be partially occluded. We calculated the performance of each block using the HSR, which naturally determined the performance of the whole image by voting. In our experiments, the image is divided into 8 (4×2) blocks, and rescaled to the size of a small (21×30 for AR database) pixel patch. In each block, After hierarchical feature selection, the dimension of Gabor-feature based image is 800, and the scale of Gabor-feature based occlusion dictionary is fixed on 20. Thus, the global dictionary in SRC is 630×1330, while the global dictionary of HSR and GSRC are 800×720. Table 4 illustrates the recognition rate and computational cost with the partition approach. The HSR with the partition achieves 99.33% in the case of sunglasses testing set and 99.00% in the case of scarves testing set with the least computational cost.

Table 4: performances of partitioning methods (SRC (p), GSRC (p) and HSR (p))

| Test set | AR sunglasses | | | AR scarves | | |
|---|---|---|---|---|---|---|
| methods | SRC(P) | GSRC(P) | HSR(P) | SRC(P) | GSRC(P) | HSR(P) |
| Rec.rate(%) | 96.00 | 99.67 | 99.33 | 93.67 | 98.67 | 99.00 |
| Time(s) | 4686.80 | 12310.00 | 1610.60 | 4693.40 | 12259.00 | 1617.00 |

For comparing difference between partitioned and no-partitioned approaches, we have visualized all results from Table3 and Table4 into Fig12 and Fig13.

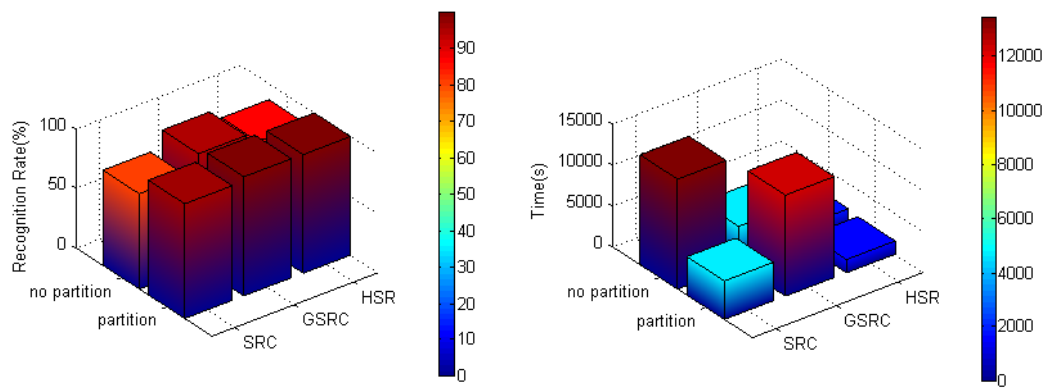

Fig12. Recognition rates and time of SRC, GSRC and HSR with no partitioning and partitioning on the sunglasses testing set

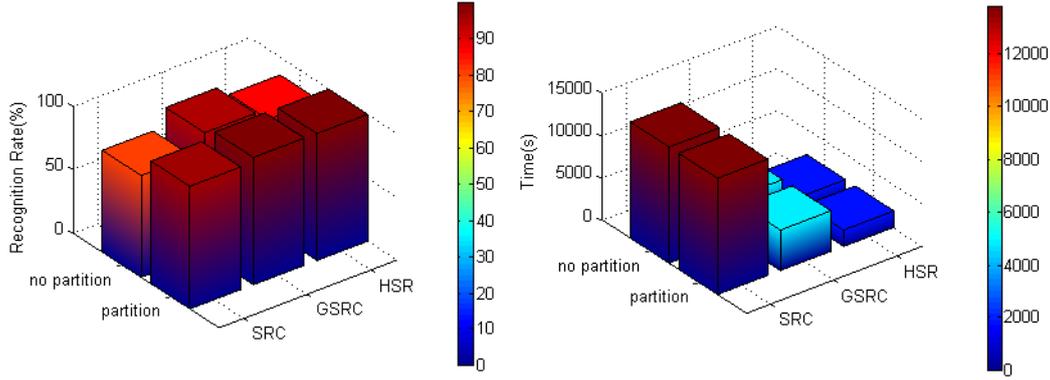

Fig13. Recognition rates and time of SRC, GSRC and HSR with no partitioning and partitioning on the scarves testing set

We can clearly see that using partitioning method, the recognition rates generally increase while the computational costs generally decrease except for GSRC. We believe that the number of sub-blocks which make wrong classifications is normally less than the number of sub-blocks that are correctly classified, which can ensure the final recognition rate. What's more, because of partitioning, the dimension and scale of occlusion dictionary are accordingly decreased, which in turn reduce the computational cost of sparse representation.

## 5  Conclusions

In this paper, we proposed a novel method for fast face recognition called $L_{1/2}$ Regularized Sparse Representation using Hierarchical Feature Selection (HSR). By employing hierarchical feature selection, we can extract the local features from image, which improves recognition rate because local features are less sensitive to the facial variation. More importantly, the global dictionary can be easily compressed in the dimension and scale by hierarchical feature selection, which speeds up the computation of sparse representation. To be more specific, it is feasible to compress the scale of Gabor-feature based occlusion dictionary via sparse coding. And high-dimensional images and global dictionary can be rapidly compressed into low-dimensional feature space via ELM-AE. By introducing $L_{1/2}$ regularized sparse representation, our method can produce sparser representation than $L_1$ regularized SRC, which in turn speeds up the face recognition. Besides, our method can also produce more robust representation than $L_1$ regularized SRC, which is more suitable to identify occluded faces such as AR sunglasses and scarves. We evaluated our method on a variety of face databases. Experimental results have demonstrated the great advantage of our method for computational cost in comparison with SRC and GSRC. Besides, we also achieve approximate or even better recognition rate. Therefore, our method has a great potential for the application of fast face recognition like real-time surveillance. Our future work will focus on two aspects. First, we will extend ELM-AE into Multi-Layer ELM-AE, which may extract more representative features in order to improve the recognition rate. Second, we will optimize the $L_{1/2}$ regularization algorithm in order to reduce the computational cost further.

**Acknowledgements**

The authors would like to thank Dr. Jun Zhou at Griffith University for helpful and excellent discussions and comments. This work is partially supported by Natural Science Foundation of China (41176076, 31202036, 51075377).